\documentclass[conference]{IEEEtran}

\makeatletter

\def\ps@IEEEtitlepagestyle{%
  \def\@oddfoot{\mycopyrightnotice}%
  \def\@evenfoot{}%
}
\def\mycopyrightnotice{%
  {\footnotesize XXX-X-XXXX-XXXX-X/XX/\$XX.00~\copyright~20XX IEEE\hfill}
  \gdef\mycopyrightnotice{}
}

\usepackage{blindtext}
\usepackage{eso-pic}
\IEEEoverridecommandlockouts
\usepackage{cite}
\usepackage{amsmath,amssymb,amsfonts}
\usepackage{algorithmic}
\usepackage{graphicx}
\usepackage{textcomp}
\usepackage{xcolor}
\usepackage{enumitem}
\usepackage{booktabs}    
\usepackage{multirow}    
\usepackage{amsmath}     
\usepackage{graphicx}
\def\BibTeX{{\rm B\kern-.05em{\sc i\kern-.025em b}\kern-.08em
    T\kern-.1667em\lower.7ex\hbox{E}\kern-.125emX}}
    
\usepackage{eso-pic}
\newcommand\AtPageUpperMyright[1]{\AtPageUpperLeft{%
 \put(\LenToUnit{0.17\paperwidth},\LenToUnit{-2cm}){%
     \parbox{0.9\textwidth}{\raggedleft\fontsize{8}{11}\selectfont #1}}%
 }}%
\newcommand{\conf}[1]{%
\AddToShipoutPictureBG*{%
\AtPageUpperMyright{#1}
}
}

\begin{document}
\title{\vspace*{1cm} Introducing Nylon Face Mask Attacks: A Dataset for Evaluating Generalised Face Presentation Attack Detection
}

\author{\IEEEauthorblockN{Manasa $^{*}$, Sushrut Patwardhan$^{\ddagger}$, Narayan Vetrekar$^{\dagger}$, Pavan Kumar$^{*}$,  R. S. Gad$^{\dagger}$, Raghavendra Ramachandra$^{\ddagger}$}
\IEEEauthorblockA{
\textit{$^{*}$Indian Institute of Information and Technology (IIIT), Dharwad, India}\\
\textit{$^{\ddagger}$Norwegian University of Science and Technology (NTNU), Gj{\o}vik, Norway}\\
\textit{$^{\dagger}$School of Physical and Applied Sciences, Goa University, Goa, India} \\
E-mail: \{raghavendra.ramachandra\} @ntnu.no}
}

\maketitle
\conf{\textit{Proc. of International Conference on Artificial Intelligence, Computer, Data Sciences and Applications (ACDSA 2026) \\ 
5-7 February 2026, Boracay-Philippines}}
\begin{abstract}
Face recognition systems are increasingly deployed across a wide range of applications, including smartphone authentication, access control, and border security. However, these systems remain vulnerable to presentation attacks (PAs), which can significantly compromise their reliability. In this work, we introduce a new dataset focused on a novel and realistic presentation attack instrument called Nylon Face Masks (NFMs), designed to simulate advanced 3D spoofing scenarios. NFMs are particularly concerning due to their elastic structure and photorealistic appearance, which enable them to closely mimic the victim's facial geometry when worn by an attacker.
To reflect real-world smartphone-based usage conditions, we collected the dataset using an iPhone 11 Pro, capturing 3,760 bona fide samples from 100 subjects and 51,281 NFM attack samples across four distinct presentation scenarios involving both humans and mannequins. We benchmark the dataset using five state-of-the-art PAD methods to evaluate their robustness under unseen attack conditions. The results demonstrate significant performance variability across methods, highlighting the challenges posed by NFMs and underscoring the importance of developing PAD techniques that generalise effectively to emerging spoofing threats.
\end{abstract}


\section{Introduction}
\label{sec:intro}
Face recognition systems (FRS) have become integral to modern biometric authentication, spanning applications from national border control to everyday access control. These systems leverage visual information to identify individuals and can operate using two-dimensional (2D) imagery, three-dimensional (3D) depth data, or multispectral imaging across various wavelengths. While advanced modalities like 3D and multispectral FRS offer enhanced robustness against presentation attacks, their deployment is often limited by hardware complexity and cost ~\cite{Survey2024PAD, ANTIL2025128992}. In contrast, 2D face recognition, based solely on RGB imagery, remains the most widely deployed due to its low-cost implementation, compatibility with conventional camera sensors, and ease of integration into existing systems ~\cite{DHS_PAD}. This has led to a proliferation of 2D FRS across multiple domains, with sources ranging from surveillance footage and ID documents to webcam and smartphone selfies. Among these, smartphone-based FRS has become particularly crucial, enabling secure and user-friendly authentication for digital banking, eKYC, device unlocking, and remote identity verification. Given the ubiquity of smartphones and the increasing reliance on mobile identity systems, ensuring the trustworthiness of 2D face recognition on these platforms is of paramount importance ~\cite{Ramachandra:2017:PAD:3058791.3038924,DHS_PAD}.
\begin{figure}[t!]
	\centering
	\includegraphics[width = .8\linewidth]{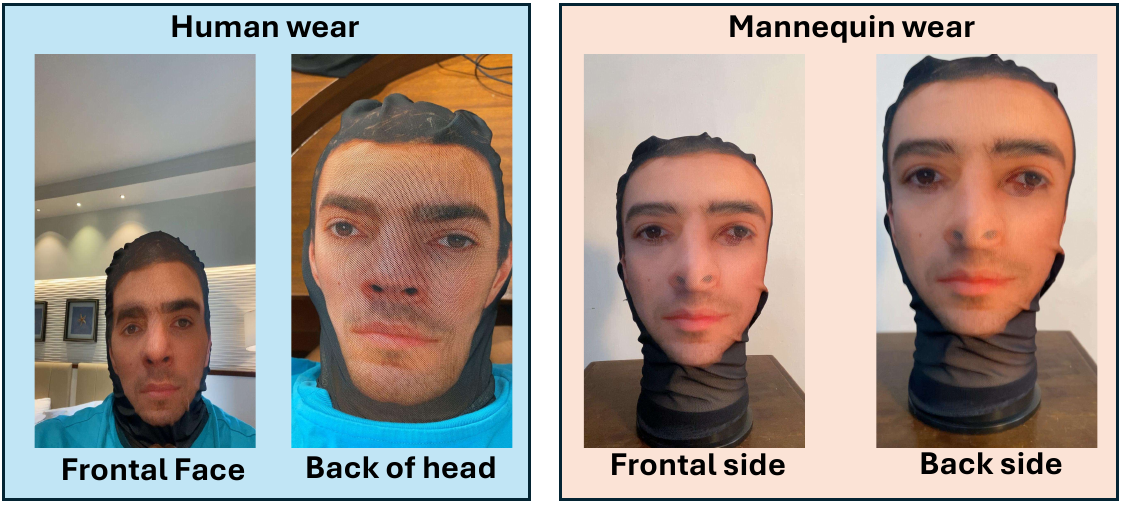}
	\caption{Illustration of the proposed Nylon Face Mask (NFM) presentation attack in different scenarios. The figure shows the NFM worn by a human subject on both the frontal face and the back of the head, as well as applied to a mannequin from the frontal and back sides. The NFM is fabricated by printing a victim's face on stretchable nylon cloth, enabling attackers to perform flexible and realistic spoofing attempts in smartphone-based face recognition systems.}
	\label{fig:Intro}
\end{figure}

However, the increased usability of facial biometrics comes with a corresponding increase in security concerns. Face Recognition Systems (FRS) are particularly vulnerable to Presentation Attacks (PAs), in which an adversary attempts to deceive the system by presenting a falsified biometric sample to the sensor. These attacks exploit the ease with which facial images can be captured or obtained, often from publicly available sources such as social media, and used to create Presentation Attack Instruments (PAIs). These PAIs range from simple printed photographs and digital screen replays to more sophisticated constructs such as photo-wrapped prints, silicone masks, and 3D-printed replicas. Additionally, the use of makeup, disguises, and facial manipulation techniques further increases the success rate of such attacks. A growing body of research~\cite{Ramachandra:2017:PAD:3058791.3038924, FaceMaskVuln, AVPAD_Survey, Survey2024PAD, Venkatesh-MADSurvey-IEEETTS-2021, DHS_PAD} has demonstrated the effectiveness of these artefacts in compromising operational FRS deployments. These findings underscore the need for robust Presentation Attack Detection (PAD) mechanisms, particularly in high-security scenarios such as border control, where failure to detect spoofing attempts could have severe implications.

Face presentation attack detection (PAD) has progressed through a wide range of algorithmic approaches. Initial methods were based on handcrafted features, with texture descriptors such as Local Binary Patterns (LBP) and its spatio-temporal extensions like LBP-TOP used to identify subtle differences in skin texture between real and spoofed faces \cite{Raghusipco, ColorLBPfacePAD, BioEvaSpoof}. Quality-based approaches further incorporated cues such as specular highlights, motion blur, colour distortions, and sensor noise to assess the authenticity of the captured face image \cite{chang2022face_Qualitry, MSUMFSD_DB, raghavendra2015presentation}. Motion-based PAD methods introduced physiological signal analysis, particularly remote photoplethysmography (rPPG), to detect natural pulse variations in facial skin regions, which are absent in presentation attacks like print or video replays \cite{liu20163d, nowara2017ppgsecure, liu2018remote}. 

The introduction of deep learning led to the widespread adoption of Convolutional Neural Networks (CNNs), which enabled automated learning of spatial and temporal features from RGB and depth data \cite{li2021diffusing, wang2022patchnet, qiao2022fgdnet, yang2014learn}. Building on this, attention-based modules and vision transformers were developed to model contextual dependencies across facial regions and to capture more complex spoofing artefacts \cite{huang2021multi, kong2022face, lucena2017transfer, george2021effectiveness_Super2, chen2019attention_Super3, wang2022face_Pix1}. More recently, anomaly detection techniques have been explored, where the model learns to represent only bona fide faces and identifies attacks as deviations from this distribution using metric learning or one-class classification strategies \cite{baweja2020anomaly, li2022one, george2020learning}. To improve generalisation to unseen environments and spoof types, domain adaptation and generalisation techniques have been introduced, including adversarial training, feature alignment, and meta-learning-based regularisation \cite{chen2021generalizable, shao2019multi, shao2020regularized, CF_FAS, cai2023rehearsal}.  These methods typically employ mechanisms such as gradient reversal, distribution matching, and episodic learning to bridge domain gaps between training and deployment scenarios. Recently, advances in multimodal foundation models have opened promising directions for PAD. Pretrained vision-language architectures and large-scale visual backbones offer strong generalisation potential, enabling few-shot or even zero-shot detection of novel attack types. These models are capable of learning rich semantic and structural representations that go beyond conventional supervised CNNs \cite{liu2024bottom, liu2024cfpl, liu2024fm, FLIP_FAS, ozgur2025foundpad}.

Despite considerable advancements in developing robust face PAD algorithms, generalisation to previously unseen attack types remains a persistent challenge. This limitation is largely due to the lack of diverse and representative datasets that adequately capture real-world variations. Many existing benchmarks offer limited diversity in terms of capture conditions, attack materials, and device-specific characteristics, particularly in scenarios involving smartphone-based face recognition systems.
To address this limitation, we introduce a new presentation attack instrument called the Nylon Face Mask (NFM). In this attack, a victim’s facial image is printed on a stretchable nylon fabric. The resulting mask can be worn directly by the attacker or placed on a mannequin and presented to the face recognition system in an attempt to gain unauthorised access. Figure~\ref{fig:Intro} illustrates several use cases of the NFM, including when worn on a human face, on the back of the head, and both the front and back of a mannequin. To the best of our knowledge, this is the first work to propose the NFM as a face presentation attack. Therefore, we evaluate the effectiveness of five state-of-the-art PAD algorithms under an unseen attack protocol to assess both the robustness of existing detection methods and the potential of this novel spoofing instrument for attack detection. The following are the main contributions of this work: 
\begin{itemize}  
\item This work introduces, for the first time, a presentation attack scenario involving 3D Nylon Face Masks in a smartphone based identity verification context.

    \item We conduct an extensive data collection campaign using five different cameras, capturing 28 unique Nylon Face Masks and 100 bona fide subjects. The data is recorded under four distinct protocols, where both human participants and mannequins wear the masks. The primary capture device is an iPhone 11 Pro, reflecting realistic mobile authentication scenarios. In total, the dataset comprises 3,760 bona fide samples and 51,218 presentation attack samples.

    \item To evaluate the robustness of existing face PAD methods against this new attack type, we benchmark five state-of-the-art approaches: 
    (a) Domain-Generalised for Unknown Attacks (DGUA)~\cite{DGUA_FAS}, 
    (b) Gradient Alignment for Cross-Domain PAD (GACD)~\cite{GACD_FAS}, 
    (c) Language-Guided Face Anti-Spoofing (FLIP)~\cite{FLIP_FAS}, 
    (d) Learnable Multi-level Frequency Decomposition (LMFD)~\cite{LMFD_FAS}, 
    (e) Causal Clues with Embedding Adaptation (CF)~\cite{CF_FAS}.

    \item The dataset will be publicly released to facilitate future research. For more details, please contact the authors of this study.
\end{itemize}

The remainder of this paper is organised as follows. Section~\ref{sec:nfm} describes the data collection process for the Nylon Face Mask (NFM) dataset and provides a summary of its statistical characteristics. Section~\ref{sec:exp} outlines the performance evaluation protocol and presents the quantitative results of five state-of-the-art PAD techniques. Finally, Section~\ref{sec:conc} concludes the paper with key findings and directions for future work.
\section{Nylon Face Mask (NFM) Dataset}
\label{sec:nfm}

In this section, we describe the data collection process for the Nylon Face Mask (NFM) dataset and present its statistical overview. The entire dataset was collected over a period of three months using an iPhone 11 Pro smartphone across multiple capture sessions. Both bona fide and attack samples were acquired under consistent capture conditions, including an indoor office environment with a mix of natural daylight and artificial lighting. This setup was intended to replicate realistic smartphone-based identity verification scenarios.

\subsection{Bona fide Data Collection:} 
\begin{figure}[htp]
	\centering
	\includegraphics[width = .9\linewidth]{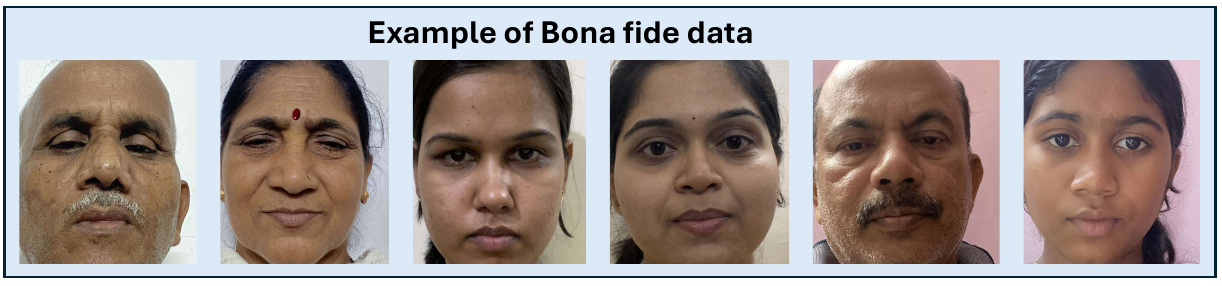}
	\caption{Examples of bona fide face images collected using the iPhone 11 Pro under realistic smartphone usage conditions. The samples were recorded in indoor office environments with varying lighting conditions to reflect natural smartphone-based identity verification scenarios.}
	\label{fig:bona}
\end{figure}

Bona fide subjects were instructed to record short selfie videos lasting between one to two seconds using the frontal camera of the iPhone 11 Pro. Each participant was asked to record at least two videos per session, captured at varying locations within the office setting. To maintain consistency with real-world ID verification use cases, participants were requested to remove spectacles and other facial accessories prior to recording.
In total, we collected data from 100 bona fide subjects (52 male and 48 female), across three distinct sessions conducted over the three-month period. The participants belong to an Asian demographic and range in age from 20 to 72 years. For use in subsequent experiments, each video was processed to extract five representative frames, resulting in a total of 3,760 bona fide face images. Figure~\ref{fig:bona} illustrates examples of the collected bona fide data.
\subsection{Attack Data Collection}
The Nylon Face Masks (NFMs) used in this study were sourced from a commercial vendor on Etsy~\cite{etsy_nfm_mask} to simulate realistic presentation attacks in smartphone-based face recognition systems. Each mask is fabricated from a stretchable nylon fabric that is both lightweight and elastic, allowing it to conform closely to the shape of a human face or mannequin. A high-resolution image of the target identity is printed directly onto the nylon surface using sublimation printing, which preserves fine facial details even when the material is stretched or curved \footnote{As the printing process was handled by a third-party vendor, detailed specifications such as the resolution or DPI of the printed face images are not disclosed and remain unknown.
}. 
The elastic nature of the fabric enables the mask to be worn securely on both the front and back of the head, enhancing the versatility and believability of the spoofing attempt. As this type of mask can be easily ordered online and manufactured at low cost, it highlights the growing feasibility of such attacks in real-world scenarios. This underscores the critical need for face PAD systems that are robust to emerging and commercially accessible attack instruments like NFMs.
In this work, we employed a total of 28 unique NFMs ordered from Etsy~\cite{etsy_nfm_mask}, representing diverse demographic attributes, including Caucasian, Asian, and African identities. Among these, 26 masks were generated for male subjects and the remaining two for female subjects.

To comprehensively evaluate the vulnerability of face recognition systems, it is important to simulate presentation attacks under diverse conditions. In this study, we use both human participants and mannequins to wear the Nylon Face Masks, thereby covering a broad spectrum of realistic attack scenarios. Human subjects introduce natural facial movements and skin texture variations, which help assess the dynamic robustness of PAD systems. In contrast, mannequins provide a stable platform to analyse the system's sensitivity to static, high-quality spoof artefacts. This dual setting enables a more thorough investigation of how different forms of presentation affect the reliability of face recognition systems in practical deployments.
\subsubsection{NFM Presentation on Human Subjects}
To simulate varied and realistic spoofing scenarios, we asked participants to wear the Nylon Face Mask (NFM) on both the frontal region of the face and the back of the head. When worn on the face, the mask naturally aligns with the underlying facial structure, preserving the three-dimensional geometry of the target identity. This closely resembles a typical high-quality 3D spoofing attempt. In contrast, wearing the mask on the back of the head introduces distortions due to hair volume, hairstyle variations, and the lack of facial contours. This experimental setup allows us to explore how facial geometry and alignment influence the effectiveness of spoofing attacks and the performance of PAD systems. It also reflects the potential for unconventional attack surfaces in real-world applications, such as surveillance or shoulder-surfing scenarios.

\begin{figure}[htp]
	\centering
	\includegraphics[width = .9\linewidth]{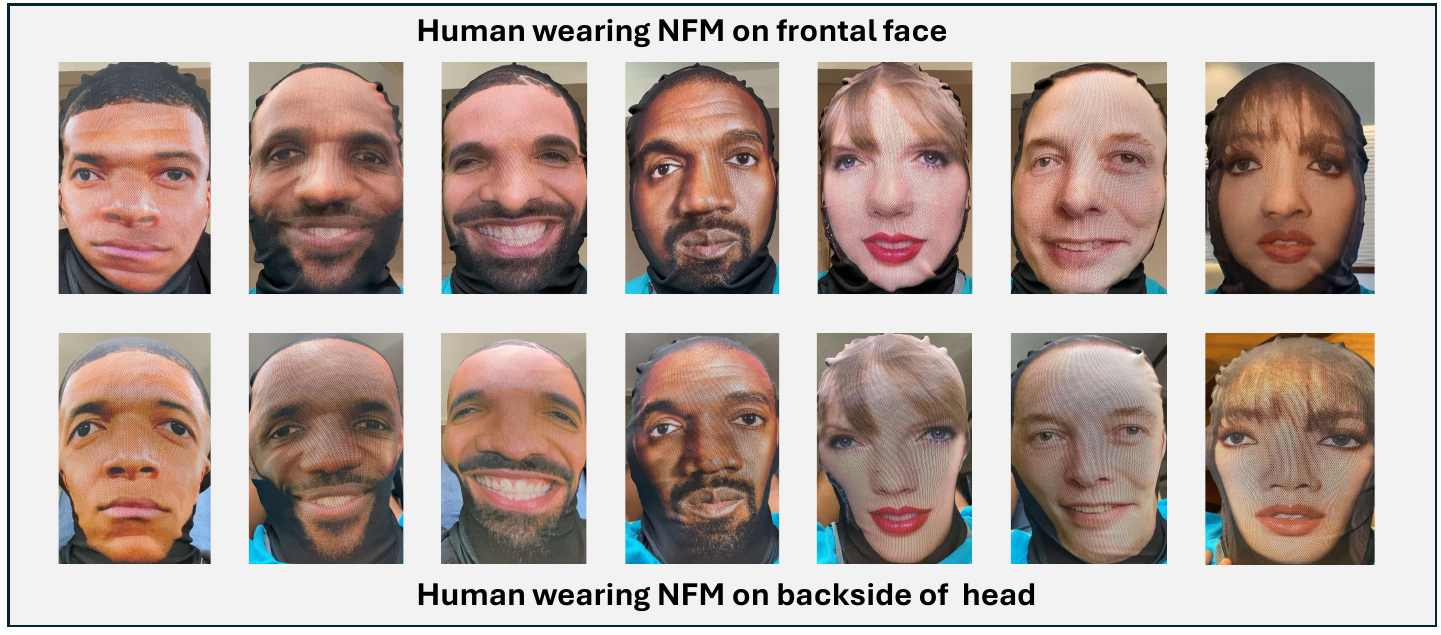}
	\caption{Examples of the Nylon Face Mask (NFM) worn by human participants. The top image shows the mask worn on the frontal face, preserving the natural three-dimensional facial structure. The bottom image shows the mask placed on the back of the head, introducing distortions due to hair and lack of facial contours.}
	\label{fig:Hum}
\end{figure}

In this scenario, data collection was carried out using 25 participants who wore the Nylon Face Mask (NFM) on both the frontal and back sides of the head. For the frontal-side recordings, participants were asked to capture short selfie videos using the front camera of an iPhone 11 Pro, following the same protocol used for bona fide data collection. Each video lasted between one to two seconds and was captured in randomly selected locations within an indoor office setting. The recordings were taken over more than 10 sessions across a period of three months to introduce natural variability. To simulate attacks from the back of the head, an assistant was employed to record the participants using the rear camera of the same device. These videos were also recorded for one to two seconds across multiple sessions during the same three-month period. For experimental use, each video was processed to extract five representative frames. This resulted in a total of 11,627 images for the frontal-face scenario and 20,794 images for the back-of-head scenario.  Figure~\ref{fig:Hum} illustrates representative examples of the NFM worn on both the front and back of the head by human subjects, highlighting the structural fidelity in frontal wear and the distortions introduced in back-of-head presentations.

\subsubsection{NFM Presentation on Mannequins}
To further evaluate the impact of surface geometry on spoofing performance, we also collected data using mannequins wearing the Nylon Face Mask (NFM) on both the frontal and back sides of the head. When the mask is worn on the front, it conforms naturally to the rigid three-dimensional shape of the mannequin face, closely simulating a structured attack scenario. In contrast, placing the mask on the back of the head introduces subtle curvature and irregularities due to the rounded surface and lack of facial features. This setup allows us to assess the sensitivity of face recognition systems to geometric inconsistencies and varying presentation angles. By using mannequins, we ensure consistent positioning and eliminate motion-related variables, making it ideal for controlled evaluation of static spoofing attacks. 

\begin{figure}[htp]
	\centering
	\includegraphics[width = .9\linewidth]{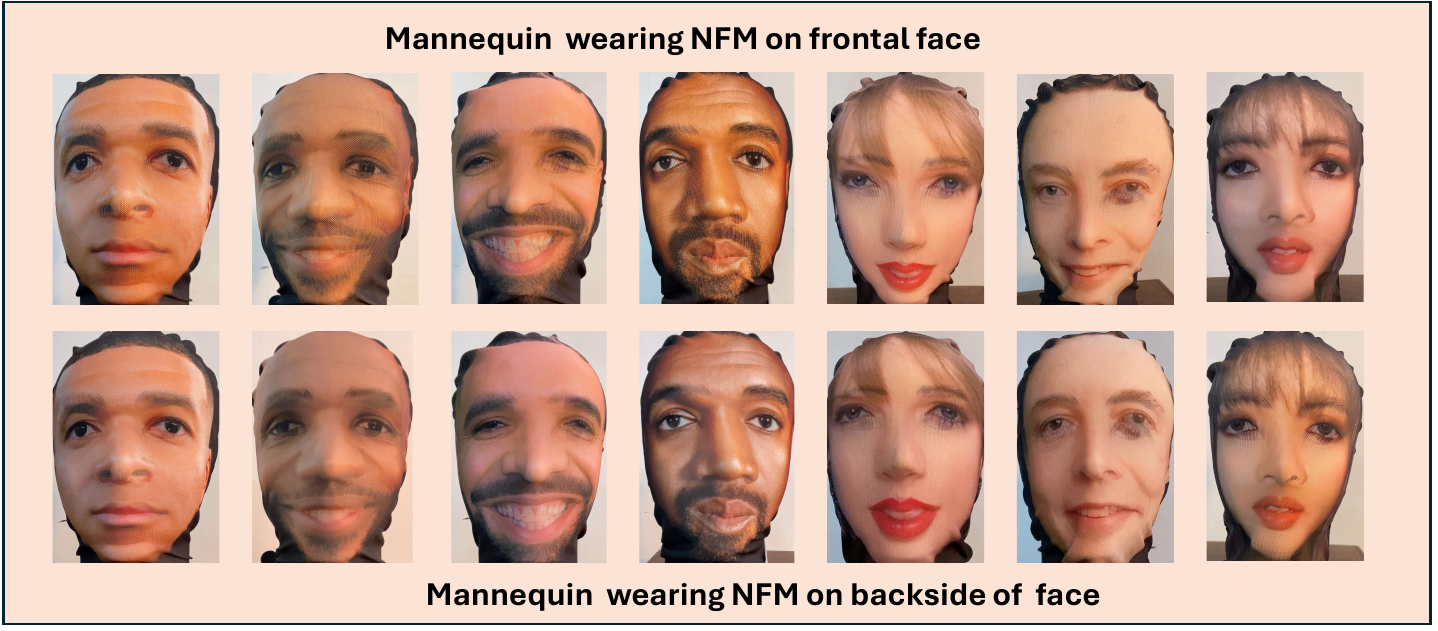}
	\caption{Examples of the Nylon Face Mask (NFM) worn by human participants. The top image shows the mask worn on the frontal face, preserving the natural three-dimensional facial structure. The bottom image shows the mask placed on the back of the head, introducing distortions due to hair and lack of facial contours.}
	\label{fig:Man}
\end{figure}

In this scenario, data was collected by placing the Nylon Face Mask (NFM) on a mannequin, both on the frontal face and the back of the head. For the frontal-side recordings, videos were captured using the front camera of an iPhone 11 Pro. In the case of back-side presentations, the rear camera of the same device was used. Videos lasting one to two seconds were recorded across multiple locations using three different mannequins in various indoor settings over a three-month period. For experimental use, each video was processed to extract five representative frames. This resulted in a total of 9,375 images for the frontal-face scenario and 9,422 images for the back-of-head scenario. Figure~\ref{fig:Man} shows representative samples of the mannequin-based data collection, illustrating the NFM worn on both the frontal face and the back of the head.
\section{Experiments and Results}
\label{sec:exp}
In this section, we present the quantitative results of the PAD methods evaluated on the newly collected Nylon Face Mask (NFM) dataset. Since NFM-based presentation attacks are introduced for the first time, we benchmark the detection performance using a set of widely adopted PAD techniques. These methods were chosen not only for their strong reported performance in prior studies but also for the diversity of their underlying approaches. They represent a broad range of methodological paradigms, including the use of novel loss functions, generative models, and vision-language foundations. In this study, we evaluate five state-of-the-art PAD techniques: Domain-Generalised for Unknown Attacks (DGUA)~\cite{DGUA_FAS}, Gradient Alignment for Cross-Domain (GACD)~\cite{GACD_FAS}, Language-Guided Face Anti-Spoofing (FLIP)~\cite{FLIP_FAS}, Learnable Multi-level Frequency Decomposition (LMFD)~\cite{LMFD_FAS}, and Causal Features with Embedding Statistics Adaptation (CF)~\cite{CF_FAS}. These methods were selected to ensure a comprehensive evaluation of detection performance under the newly introduced NFM attack scenario.
The detection performance of all PAD algorithms is reported using standard ISO/IEC 30107-3 \cite{ISO-IEC-30107-3-PAD-metrics-170227} metrics. Specifically, the Attack Presentation Classification Error Rate (APCER) refers to the proportion of attack samples that are incorrectly classified as bona fide. Conversely, the Bona Fide Presentation Classification Error Rate (BPCER) measures the proportion of bona fide samples that are wrongly classified as attack presentations \cite{ISO-IEC-30107-3-PAD-metrics-170227}. In addition, we also report the quantitative results using the Detection Equal Error Rate (D-EER), which corresponds to the point at which the BPCER equals the APCER.

\subsection{Performance Evaluation Protocol}
We adopt an unseen attack detection protocol to evaluate the generalisability of state-of-the-art PAD models. In this work, we do not perform any training but instead use pre-trained models that were originally trained on four public datasets: OULU-NPU~\cite{OULUNPU_DB}, MSU-MFSD~\cite{MSUMFSD_DB}, Replay-Attack~\cite{ReplayAttcak_DB}, and CelebA-Spoof~\cite{CelebA-Spoof_DB}. Each model follows a leave-one-dataset-out training scheme, where three datasets are used for training and one is held out. These trained models are directly applied to our NFM dataset, which acts as a novel and unseen attack source. This setup reflects a realistic deployment scenario and enables robust analysis of the PAD models under cross-dataset conditions. Final performance is reported as the mean and standard deviation over four such runs.
\subsection{Results and discussion}
\begin{table}[htp]
\centering
\caption{Quantitative results on human-subject-based NFM attacks. Results are reported as mean $\pm$ standard deviation for Detection Equal Error Rate (D-EER), and BPCER at two fixed values of APCER (5\% and 10\%). Best values in each column are shown in \textbf{bold}.}
\label{tab:human_nfm_results}
\resizebox{0.99\columnwidth}{!}{%
\begin{tabular}{@{}llccc@{}}
\toprule
\textbf{PAI Type} & \textbf{PAD Algorithm} & \textbf{D-EER (\%)} & \textbf{BPCER @ 5\% APCER} & \textbf{BPCER @ 10\% APCER} \\
\midrule
\multirow{5}{*}{Front side} 
& DGUA~\cite{DGUA_FAS} & $3.39 \pm 2.77$ & $3.41 \pm 2.83$ & $2.95 \pm 2.34$ \\
& GACD~\cite{GACD_FAS} & $17.41 \pm 20.82$ & $32.62 \pm 37.83$ & $32.49 \pm 37.65$ \\
& FLIP~\cite{FLIP_FAS} & $9.07 \pm 2.31$ & $10.41 \pm 3.12$ & $8.83 \pm 2.85$ \\
& LMFD~\cite{LMFD_FAS} & $12.57 \pm 0.91$ & $16.52 \pm 2.95$ & $13.56 \pm 1.64$ \\
& CF~\cite{CF_FAS}     & \textbf{1.95 $\pm$ 0.94} & \textbf{1.94 $\pm$ 0.51} & \textbf{1.78 $\pm$ 0.41} \\
\midrule
\multirow{5}{*}{Back side} 
& DGUA~\cite{DGUA_FAS} & \textbf{4.43 $\pm$ 0.82} & \textbf{3.44 $\pm$ 1.74} & \textbf{1.17 $\pm$ 0.92} \\
& GACD~\cite{GACD_FAS} & $28.40 \pm 32.13$ & $35.70 \pm 36.73$ & $32.83 \pm 38.13$ \\
& FLIP~\cite{FLIP_FAS} & $5.89 \pm 1.58$ & $6.65 \pm 3.22$ & $4.13 \pm 2.57$ \\
& LMFD~\cite{LMFD_FAS} & $9.19 \pm 1.29$ & $12.88 \pm 2.91$ & $8.47 \pm 2.48$ \\
& CF~\cite{CF_FAS}     & $6.28 \pm 2.96$ & $7.75 \pm 3.66$ & $4.91 \pm 3.04$ \\
\bottomrule
\end{tabular}
}
\end{table}

\begin{figure*}[!htb]
\minipage{0.5\textwidth}
  \includegraphics[width=\linewidth]{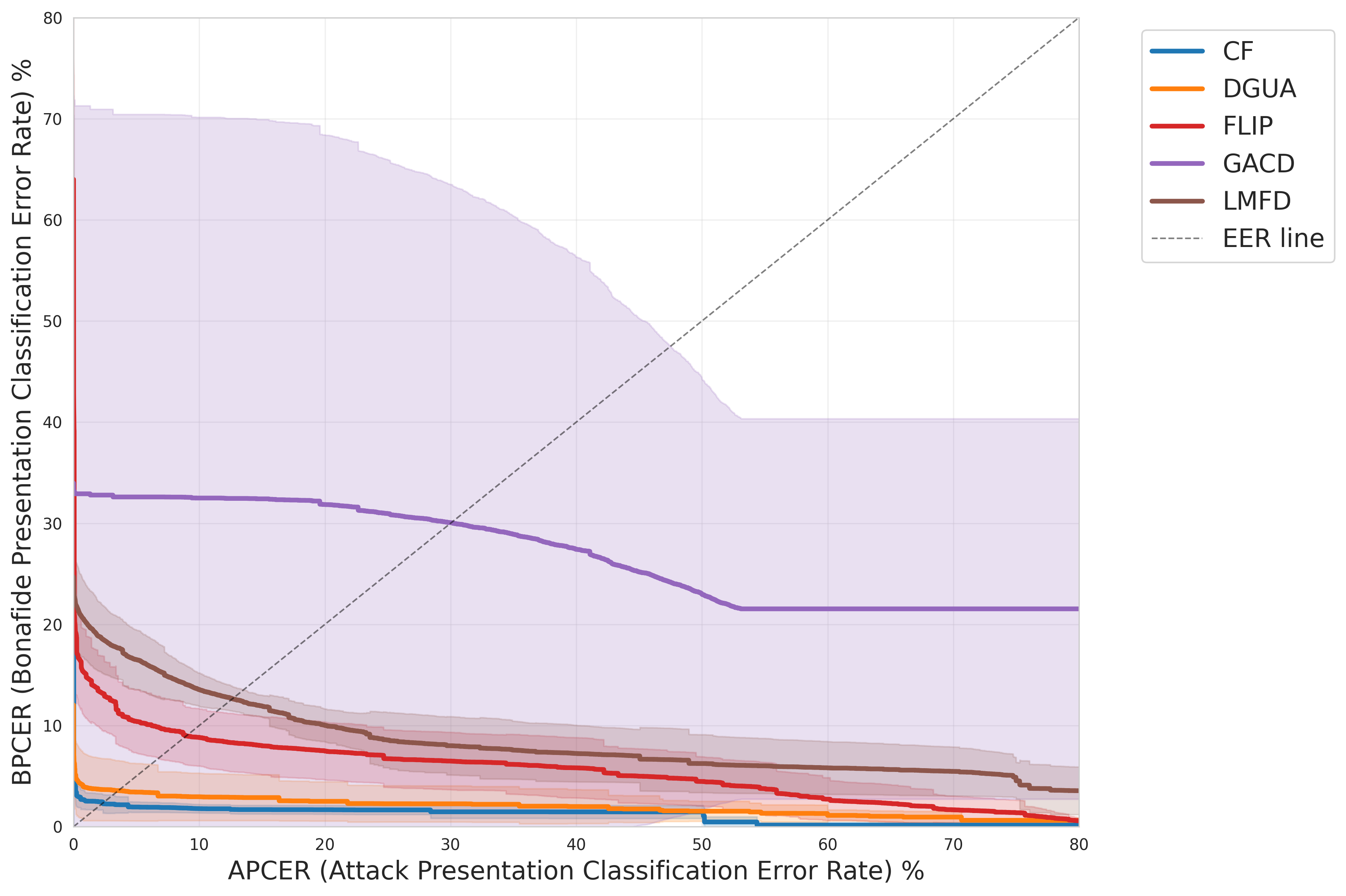}
  \caption{DET curves for Human-Front side}\label{fig:awesome_image1}
\endminipage\hfill
\minipage{0.5\textwidth}
  \includegraphics[width=\linewidth]{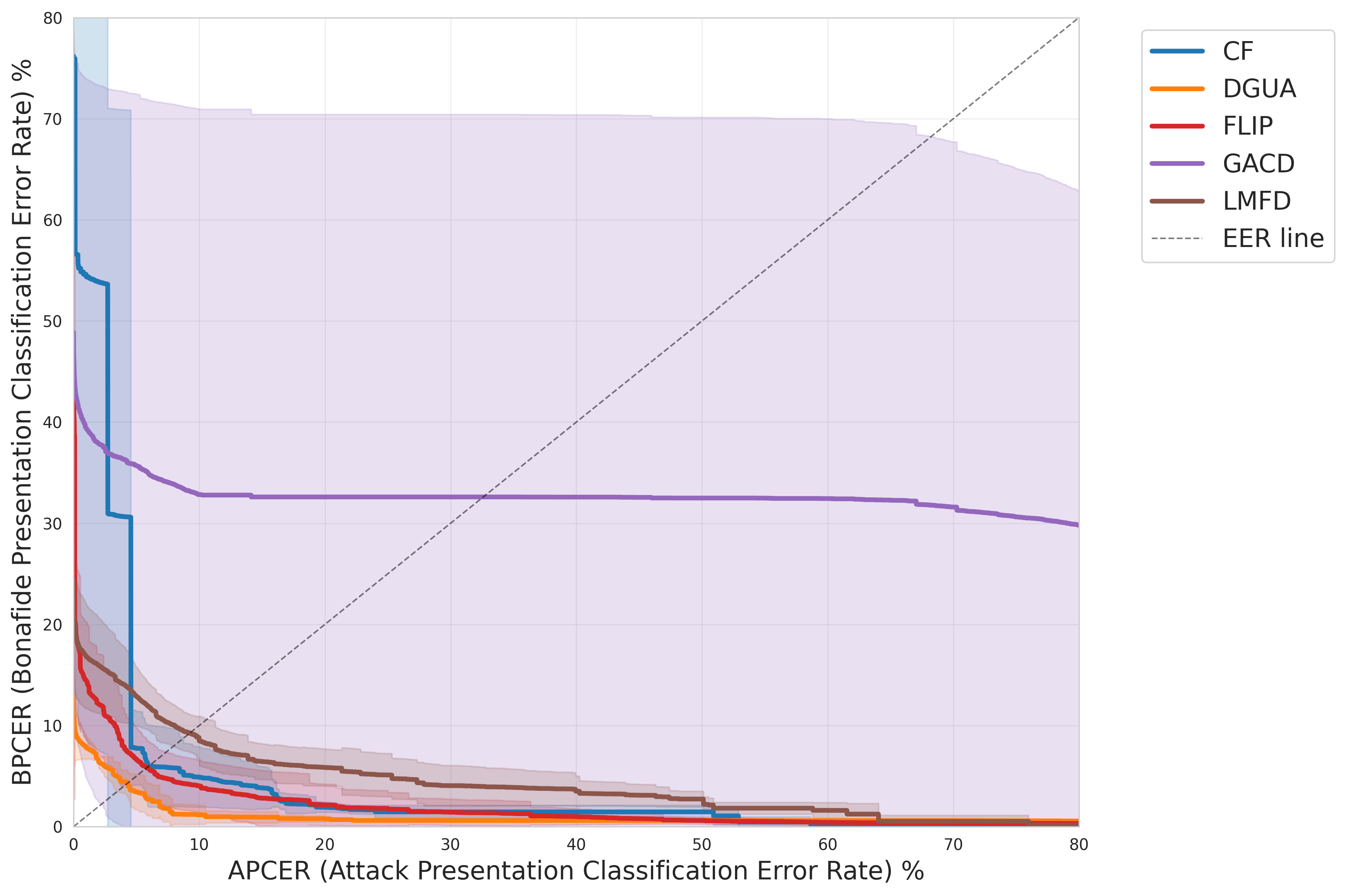}
  \caption{DET curves for Human-Back side}\label{fig:awesome_image2}
\endminipage
\label{fig:DET1}
\end{figure*}

The results in Table~\ref{tab:human_nfm_results} reveal that the CF~\cite{CF_FAS} method achieves the best overall performance for NFM attacks presented on the front side of human subjects, with the lowest D-EER of $1.95\%$ and consistently superior BPCER values across both APCER thresholds. This can be attributed to CF's ability to model causal dependencies and adaptively regulate embedding statistics, making it more robust to texture inconsistencies and lighting distortions caused by the elastic nylon material. For back-side NFM attacks, DGUA~\cite{DGUA_FAS} shows the best detection performance, achieving a D-EER of $4.43\%$. Its domain generalisation framework, which learns invariant features across datasets, appears effective in handling the spatial deformations and occlusions introduced by the rear-side presentations.

Interestingly, front-side attacks exhibit lower D-EERs than back-side attacks across most PAD methods. Although the front-facing masks closely resemble natural facial geometry, they tend to introduce uniform artefacts such as facial blurring, stretching, and reduced high-frequency texture, which PAD models are better able to learn and detect. On the other hand, when the mask is worn on the back of the head, variations due to hairstyle, shape asymmetry, and camera positioning introduce greater intra-class variability, leading to a degradation in detection performance. Figure~\ref{fig:DET1} shows the DET curves with mean and standard deviation for the human-presented NFM experiments, further illustrating the separation between bona fide and spoof samples under both scenarios.

Overall, these results suggest that existing PAD methods are capable of detecting NFMs with promising accuracy, especially when spoof artefacts are consistent and learnable. However, the variation in performance across presentation styles highlights the importance of dataset diversity and representation in training PAD models. Figure~\ref{fig:DET1} shows the DET curves with mean and standard deviation for the human-presented NFM experiments, further illustrating the separation between bona fide and spoof samples under both scenarios.

\begin{table}[htp]
\centering
\caption{Quantitative results on mannequin-based NFM attacks. Results are reported as mean $\pm$ standard deviation for Detection Equal Error Rate (D-EER), and BPCER at two fixed APCER values (5\% and 10\%). Best values in each column are shown in \textbf{bold}.}
\label{tab:mannequin_nfm_results}\resizebox{0.99\columnwidth}{!}{%
\begin{tabular}{@{}llccc@{}}
\toprule
\textbf{PAI Type} & \textbf{PAD Algorithm} & \textbf{D-EER (\%)} & \textbf{BPCER @ 5\% APCER} & \textbf{BPCER @ 10\% APCER} \\
\midrule
\multirow{5}{*}{Front side} 
& DGUA~\cite{DGUA_FAS} & \textbf{4.15 $\pm$ 2.37} & \textbf{3.84 $\pm$ 2.49} & \textbf{3.45 $\pm$ 2.62} \\
& GACD~\cite{GACD_FAS} & $20.00 \pm 24.85$ & $32.45 \pm 37.58$ & $32.37 \pm 37.44$ \\
& FLIP~\cite{FLIP_FAS} & $15.67 \pm 6.37$ & $41.68 \pm 34.14$ & $35.05 \pm 37.52$ \\
& LMFD~\cite{LMFD_FAS} & $12.51 \pm 1.85$ & $16.12 \pm 3.15$ & $13.29 \pm 2.56$ \\
& CF~\cite{CF_FAS}     & $6.66 \pm 1.48$ & $29.49 \pm 40.70$ & $4.82 \pm 1.14$ \\
\midrule
\multirow{5}{*}{Back side} 
& DGUA~\cite{DGUA_FAS} & \textbf{3.39 $\pm$ 2.29} & \textbf{3.11 $\pm$ 2.42} & \textbf{2.51 $\pm$ 2.53} \\
& GACD~\cite{GACD_FAS} & $20.18 \pm 23.66$ & $32.89 \pm 37.36$ & $32.29 \pm 37.31$ \\
& FLIP~\cite{FLIP_FAS} & $14.87 \pm 5.04$ & $32.89 \pm 37.36$ & $18.04 \pm 6.74$ \\
& LMFD~\cite{LMFD_FAS} & $11.12 \pm 1.45$ & $13.86 \pm 1.51$ & $11.51 \pm 1.89$ \\
& CF~\cite{CF_FAS}     & $6.23 \pm 1.03$ & $6.20 \pm 0.82$ & $4.51 \pm 1.12$ \\
\bottomrule
\end{tabular}
}
\end{table}

Table~\ref{tab:mannequin_nfm_results} summarises the PAD performance for NFM attacks presented using mannequins. The DGUA~\cite{DGUA_FAS} method consistently delivers the best results for both front and back-side scenarios, with the lowest D-EER of $3.39\%$ for the back-side attacks. Its cross-domain learning framework appears to be particularly effective in extracting stable features under controlled lighting and texture, which are more consistent when using mannequins. Notably, FLIP~\cite{FLIP_FAS} and GACD~\cite{GACD_FAS} show substantial performance degradation, suggesting that methods relying on large-scale pretraining or gradient alignment may be more sensitive to low-frequency surface texture or lighting artefacts introduced by the mask material and mannequin contours.

Compared to human-based NFM attacks, mannequin-based attacks show a slightly more uniform performance across PAD algorithms. While CF~\cite{CF_FAS} dominated human presentations, its performance here is less stable, likely due to reduced dynamic facial cues when masks are worn on inanimate heads. The lower D-EER values for back-side mannequin presentations suggest that geometric distortion and simplified background context may contribute to more consistent spoof cues, which some algorithms can exploit. These findings highlight that the absence of natural facial dynamics in mannequins alters the spoofing artefact distribution, necessitating PAD systems that adapt to both texture and motion-based discrepancies in a unified manner.
\section{Conclusion}
\label{sec:conc}

Presentation attack instruments (PAIs) continue to pose a significant threat to the reliability and trustworthiness of face recognition systems. The development of robust presentation attack detection (PAD) techniques critically depends on the availability of diverse and realistic attack scenarios. Introducing novel PAIs is therefore essential for driving progress in PAD research and ensuring that systems remain resilient against evolving spoofing strategies.
In this work, we presented a new PAI based on Nylon Face Masks (NFMs) and introduced a comprehensive smartphone-based dataset simulating realistic attack conditions. The dataset includes 3,760 bona fide samples from 100 subjects and 51,218 NFM attack samples collected under four different presentation scenarios involving both human participants and mannequins. To our knowledge, this is the first dataset to systematically evaluate the impact of NFM-based attacks on PAD performance.
We benchmarked five state-of-the-art PAD methods under an unseen attack protocol and observed that while some techniques demonstrated strong detection performance on frontal-face scenarios, others struggled under more complex back-side or mannequin-based attacks. These results highlight the importance of evaluating PAD systems against previously unseen and physically plausible attacks such as NFMs. Despite achieving reasonable PAD performance, the NFM attack remains a significant threat due to its realism, low cost, and ease of replication, underscoring the need for PAD systems to remain resilient against emerging artefacts.
In future work, we plan to expand the dataset with additional PAI variants and capture conditions, and to explore the use of multimodal learning and vision-language foundation models to enhance generalisation across attack domains.



\scriptsize
\bibliographystyle{IEEEtran}
\bibliography{2DFacePAD}

\end{document}